\documentclass[11pt,a4paper]{article}
\usepackage{authblk}
\usepackage[hyperref]{acl2020}
\usepackage{times}
\usepackage{latexsym}

\usepackage{microtype}
\usepackage{url}
\usepackage{adjustbox}
\usepackage{multirow}
\usepackage{tabularx}
\usepackage{textcomp}
\usepackage{soul}
\usepackage{tcolorbox}
 \usepackage{amssymb}
\usepackage{hyperref}
\aclfinalcopy 
\def\aclpaperid{1575} 

\title{Controlled Crowdsourcing for High-Quality QA-SRL Annotation}

\makeatletter
\renewcommand\AB@authnote[1]{\rlap{\textsuperscript{\normalfont#1}}}

\makeatother

\author[1]{\bf Paul Roit}
\author[1]{\bf Ayal Klein}
\author[1]{\bf Daniela Stepanov}
\author[3]{\bf Jonathan Mamou}

\author[2]{\\ \bf Julian Michael}
\author[2,4]{\bf Gabriel Stanovsky}
\author[2,5]{\bf \, Luke Zettlemoyer}
\author[1]{\bf Ido Dagan}
\affil[1]{Department of Computer Science, Bar-Ilan University, Ramat-Gan, Israel}
\affil[2]{Paul G. Allen School of Computer Science \& Engineering,}
\affil[ ]{\vspace{0em} University of Washington, Seattle, USA}
\affil[3]{Intel AI Lab, Israel}
\affil[4]{Allen Institute for AI, Seattle, USA}
\affil[5]{Facebook AI Research}
\affil[  ]{\vspace{-1em}}
\affil[  ]{ \small \tt \{plroit, ayal.s.klein, daniela.stepanov\}@gmail.com jonathan.mamou@intel.com}
\affil[  ]{\small \tt \{julianjm, gabis, lsz\}@cs.washington.edu dagan@cs.biu.ac.il} 
\date{}
\begin{document}
\maketitle

\begin{abstract}
Question-answer driven Semantic Role Labeling (QA-SRL) was proposed as an attractive open and natural flavour of SRL, potentially attainable from laymen.
Recently, a large-scale crowdsourced QA-SRL corpus and a trained parser were released. 
Trying to replicate the QA-SRL annotation for new texts, we found that the resulting annotations were lacking in quality, particularly in coverage, making them insufficient for further research and evaluation. 
In this paper, we present an improved crowdsourcing protocol for complex semantic annotation, involving worker selection and training, and a data consolidation phase. 
Applying this protocol to QA-SRL yielded high-quality annotation with drastically higher coverage, producing a new gold evaluation dataset. 
We believe that our annotation protocol and gold standard will facilitate future replicable research of natural semantic annotations.

\end{abstract}

\section{Introduction}

Semantic Role Labeling (SRL) provides explicit annotation of predicate-argument relations.
Common SRL schemes, particularly PropBank \cite{martha2005propbank} and FrameNet \cite{Baker:1998:BFP:980451.980860}, rely on predefined role inventories and extensive predicate lexicons. 
Consequently, SRL annotation of new texts requires substantial efforts involving expert annotation, and possibly lexicon extension, limiting scalability. 

Aiming to address these limitations, Question-Answer driven Semantic Role Labeling (QA-SRL) \cite{He2015qasrl} labels each predicate-argument relationship with a question-answer pair, where natural language questions represent semantic roles, and answers correspond to arguments (see Table \ref{table:examples}).
This approach follows the colloquial perception of semantic roles as answering questions about the predicate (``Who did What to Whom, When, Where and How", with, e.g., ``Who" corresponding to the \textit{agent} role).

QA-SRL carries two attractive promises. First, using a question-answer format makes the annotation task intuitive and easily attainable by laymen, as it does not depend on linguistic resources (e.g. role lexicons), thus facilitating greater annotation scalability. 
Second, by relying on intuitive human comprehension, these annotations elicit a richer argument set, including valuable \textit{implicit} semantic arguments not manifested in syntactic structure
(highlighted in Table \ref{table:examples}). 
The importance of implicit arguments has been recognized in the literature \cite{cheng2018implicit, do2017improving_implicit, gerber2012semantic_implicit}, yet they are mostly overlooked by common SRL formalisms and tools.

Overall, QA-SRL largely subsumes predicate-argument information  captured by traditional SRL schemes, which were shown beneficial for complex downstream tasks, such as dialog modeling \cite{chen2013unsupervised_srlInDialog}, machine comprehension \cite{wang2015machine} and cross-document coreference \cite{barhom2019revisiting}. 
At the same time, it contains richer information, and is easier to understand and collect.   
Similarly to SRL, one can utilize QA-SRL both as a source of semantic supervision, in order to achieve better implicit neural NLU models, as done recently by \citet{QuASE_HeNgRo20}, as well as an explicit semantic structure for downstream use, e.g. for producing Open Information Extraction propositions \cite{stanovsky2016creating}.\footnote{Indeed, making direct use of QA-SRL role questions might seem more challenging than with categorical semantic roles, as in traditional SRL. In practice, however, when a model embeds QA-SRL questions in context, we would expect similar embeddings for semantically similar questions. 
These embeddings may be leveraged downstream in the same way as embeddings of traditional categorical semantic roles.}

Previous attempts to annotate QA-SRL initially involved trained annotators \cite{He2015qasrl} but later resorted to crowdsourcing \cite {fitz2018qasrl} for scalability.
Naturally, employing crowd workers is challenging when annotating fairly demanding structures like SRL.
As \citet{fitz2018qasrl} acknowledge, the main shortage of their large-scale dataset is limited recall, which we estimate to be in the lower 70s (see \S \ref{sec:dataset_quality_analyses}). Unfortunately, such low recall in gold standard datasets hinders proper research and evaluation, undermining the current viability of the QA-SRL paradigm.

Aiming to enable future QA-SRL research, 
we present a generic \emph{controlled crowdsourcing} annotation protocol and apply it to QA-SRL. 
Our process addresses worker quality by performing short yet efficient annotator screening and training.
To boost coverage, we employ two independent workers per task, while an additional worker resolves inconsistencies, similar to conventional expert annotation. 
These steps combined yield 25\% more roles than \citet{fitz2018qasrl}, without sacrificing precision and at a comparable cost per verb. 
This gain is especially notable for implicit arguments,
which we show in a comparison to PropBank \cite{martha2005propbank}. 
Overall, we show that our annotation protocol and dataset are of high quality and coverage, enabling subsequent QA-SRL research.

To foster such research, including easy production of additional QA-SRL datasets, we release our annotation protocol, software and guidelines along with a high-quality dataset for QA-SRL evaluation (dev and test).\footnote{\url{https://github.com/plroit/qasrl-gs}}
We also re-evaluate the existing parser \cite{fitz2018qasrl} against our test set, setting the baseline for future developments. 
Finally, we propose that our systematic and replicable controlled crowdsourcing protocol could also be effective for other complex annotation tasks.\footnote{A previous preprint version of this paper can be found at \url{https://arxiv.org/abs/1911.03243}. }

\begin{table}[t!]
\centering
\resizebox{\columnwidth}{!}{
\begin{tabular}{l{c}r{c}}
\multicolumn{2}{p{11cm}}{Around 47 people could be \textbf{arrested}, including \hl{the councillor}. } \\ \hline
(1) Who might be \textbf{arrested}? & 47 people \textbar\hspace{0pt} the councillor \\
\hline\hline
\multicolumn{2}{p{10.5cm}}{Perry called for the DA’s resignation, and when \hl{she did not resign}, \textbf{cut} funding to a program she ran.} \\ \hline
(2) Why was something \textbf{cut} by someone? & she did not resign \\
(3) Who \textbf{cut} something? & Perry
\end{tabular}
}
\caption{QA-SRL examples. The bar (\textbar) separates multiple answers. Implicit arguments are highlighted.}
\label{table:examples}

\end{table}
\begin{table}[ht]
    \centering
    \resizebox{\columnwidth}{!}{
    \begin{tabular}{c|c|c|c|c|c|c c}
         \textbf{WH} & \textbf{AUX} & \textbf{SUBJ} & \textbf{TGT} & \textbf{OBJ} & \textbf{PREP} & \textbf{MISC} & ?\\ \hline\hline  
         Why& was& something & cut & & by & someone & ? \\ \hline 
         Why & did & someone & cut & something & & & ? \\ \hline
         Who& might& & be arrested & & & & ?
         
    \end{tabular}}
    \caption{Examples for the question template corresponding to the 7 slots. First two examples are semantically equivalent.}
    \label{tab:question_slots}
\end{table}

\section{Background --- QA-SRL}\label{sec:background}

\paragraph{Specifications}

In QA-SRL, a role question adheres to a 7-slot template, with slots corresponding to a WH-word, the verb, auxiliaries, argument placeholders (SUBJ, OBJ), and prepositions, where some slots are optional \cite{He2015qasrl}, as exemplified in Table~\ref{tab:question_slots}.
Such a question captures its corresponding semantic role with a natural, easily understood expression. 
All answers to the question are then considered as the set of arguments associated with that role, capturing both traditional explicit arguments and implicit ones.




\paragraph{Corpora}
The original {\sf 2015} QA-SRL dataset \cite{He2015qasrl} was annotated by hired non-expert workers after completing a short training procedure. 
They annotated 7.8K verbs, reporting an average of 2.4 QA pairs per verb.
Even though multiple annotators were shown to produce greater coverage, their released dataset was produced by a single annotator per verb. 

In subsequent work, \citet{fitz2018qasrl} employed untrained crowd workers to construct a large-scale corpus ({\sf 2018}) and used it to train a parser.
In their protocol, a single worker (``generator") annotated a set of questions along with their answers. Two additional workers (``validators") validated each question and, in the valid case, independently annotated their own answers.
In total, 133K verbs were annotated with 2.0 QA pairs per verb on average.

In a subset of the corpus (10\%) reserved for parser evaluation, verbs were densely validated by 5 workers (termed the {\sf Dense} set).\footnote{\citet{fitz2018qasrl} also produced an expanded version of their dataset, incorporating questions that were automatically generated by their parser and then validated by crowd workers.
While this may achieve higher recall, using model-generated data biases the evaluation with respect to existing models and is not suitable for evaluation datasets. 
For that reason, in our work we consider only the non-expanded version of the \textsf{Dense} set.}
Yet, adding validators accounts only for precision errors in question annotation, while role coverage solely relies upon the output of the single generator. For this reason, both the {\sf 2015} and {\sf 2018} datasets struggle with coverage. 

Also, while traditional SRL annotations contain a single authoritative and non-redundant annotation (i.e., a single role and span for each argument), the {\sf 2018} dataset provides raw annotations from all annotators. 
These include many \emph{redundant} overlapping argument spans, without settling on consolidation procedures to provide a single gold reference, which complicates models' evaluation.

These limitations of the current QA-SRL datasets impede their utility for future research and evaluation.
Next, we describe our method for creating a viable high quality QA-SRL dataset.

\section{Annotation and Evaluation Methods}
\label{sec:annotation_methodology}

\subsection{Controlled Crowdsourcing Methodology}
\label{sec:crowdsourcing_methodology}
\paragraph{Screening and Training}
We first release a preliminary crowd-wide annotation round, and then contact workers who exhibit reasonable performance. 
They are asked to review our short guidelines,\footnote{Publicly available in our repository.} which highlight a few subtle aspects, and then annotate two qualification rounds, of 15 predicates each. 
Each round is followed by extensive feedback via email, pointing at  errors and missed arguments, identified by automatic comparison to expert annotation. Total worker effort for the training phase is about 2 hours, and is fully compensated, while requiring about half an hour of an in-house trainer time per participating worker. We trained 30 participants, eventually selecting 11 well-performing ones.

\paragraph{Annotation} 
\label{sec:consolidation}
We reuse and extend the annotation machinery of \citeauthor{fitz2018qasrl} over Amazon's Mechanical Turk.
First, two workers independently generate questions about a verb, and highlight answer spans in the sentence.
Then, a third worker reviews and \textit{consolidates} their annotations based on targeted guidelines, producing the gold standard data. At this step, the worker validates questions, merges, splits or modifies answers for the same role, and removes redundant questions.\footnote{Notice that while the validator from \citet{fitz2018qasrl} viewed only the questions of a single generator, our consolidator views two full QA sets, promoting higher coverage.}
Table~\ref{tab:consolidation} depicts examples from the consolidation task.
We monitor the annotation process by sampling (1\%) and reviewing.
\paragraph{Data \& Cost}
We annotated a sample of the {\sf Dense} evaluation set, comprising of 1000 sentences from each of the Wikinews and Wikipedia domains, equally split to dev and test.
Annotators are paid 5\textcent\hspace{0pt} per predicate for QA generation, with an additional bonus for every question beyond the first two. 
The consolidator is rewarded 5\textcent\hspace{0pt} per verb and 3\textcent\hspace{0pt} per question. 
Per predicate, on average, our cost is 54.2\textcent\hspace{0pt}, yielding 2.9 roles, compared to reported 2.3 valid roles with approximately 51\textcent\hspace{0pt} per predicate for the {\sf Dense} annotation protocol.
\begin{table}[t!]
\resizebox{\columnwidth}{!}{
\begin{tabular}{lll}

 A1: & Who identified something?&  The U.S. Geological Survey (USGS)\\
 A2: & Who identified something?&  The U.S. Geological Survey\\\hline\hline
C: & Who identified something & \textbf{The U.S. Geological Survey} \textbar{} \textbf{USGS}\\ \hline\hline 
A1: &What might contain something? & that basin\\
A2: & What contains something? & that basin\\
\hline\hline
C: & \textbf{What might contain something?} & that basin
\end{tabular}}
\caption{Example annotations for the consolidation task. A1 and A2 refer to question-answer pairs of the original annotators, while C refers to the consolidator-selected question and corrected answers. }
\label{tab:consolidation}
\end{table}
\subsection{Evaluation Metrics}
\label{sec:eval_metrics}
Evaluation in QA-SRL involves, for each verb, aligning its predicted argument spans to a reference set of arguments, and evaluating question equivalence, i.e., whether predicted and gold questions for aligned spans correspond to the same semantic role. 
Since detecting question equivalence is still an open challenge, we propose both unlabeled and labeled evaluation metrics.
The described procedure is used to evaluate both the crowd-workers' annotations (\S\ref{sec:dataset_quality_analyses}) and the QA-SRL parser (\S\ref{sec:baseline_parser}).

\paragraph{Unlabeled Argument Detection ({\sf UA})} 
Inspired by the method presented in \cite{fitz2018qasrl}, argument spans are matched using a token-based matching criterion of intersection over union \textsc{(IOU)} $\geq 0.5$. To credit each argument only once, we employ maximal bipartite matching\footnote{The previous approach aligned arguments to roles. We measure argument detection, whereas \citet{fitz2018qasrl} measure role detection.} between the two sets of arguments, drawing an edge for each pair that passes the above mentioned criterion. The resulting maximal matching determines the true-positive set, while remaining non-aligned arguments become false positives or false negatives.
\paragraph{Labeled Argument Detection ({\sf LA}) } 
All aligned arguments from the previous step are inspected for label equivalence, similar to the joint evaluation reported in \citet{fitz2018qasrl}. 
There may be many correct questions for a role. 
For example, \textit{What was given to someone?} and \textit{What has been given by someone?} both refer to the same semantic role but diverge in grammatical tense
and argument place holders.
Aiming to avoid judging non-equivalent roles as equivalent, we propose \textsc{Strict-Match} to be an equivalence on the following template slots: \textbf{WH}, \textbf{SUBJ}, \textbf{OBJ}, as well as on negation, voice, and modality\footnote{Presence of factuality-changing modal verbs such as \textit{should}, \textit{might} and \textit{can}.} extracted from the question. 
Final reported numbers on labelled argument detection rates are based on bipartite aligned arguments passing      \textsc{Strict-Match}.
As this matching criterion significantly underestimates question equivalence, we later manually assess the actual rate of correct role equivalences.  
\paragraph{Evaluating Redundant Annotations} We extend our metric for evaluating manual or automatic redundant annotations, exhibited in the {\sf Dense} dataset (\S\ref{sec:background}) as well as the output of the \citet{fitz2018qasrl} parser  (\S\ref{sec:baseline_parser}).
 To that end, we ignore redundant true-positives, and collapse false-positive errors (see Appendix for details).

\section{Dataset Quality Analysis}
\label{sec:dataset_quality_analyses}

\paragraph{Inter-Annotator Agreement (IAA)}
\label{sec:IAA}

To estimate dataset consistency across different annotations, we measure F1 using our \textsf{UA} metric.
10 individual worker-vs-worker experiments  yield \textbf{79.8 } F1 agreement over 150 predicates, indicating high consistency across our annotators, in line with agreement rates in other structured semantic annotations, e.g. \citet{abend2013ucca}.
Overall consistency of the dataset is assessed by measuring agreement between different \textit{consolidated} annotations, obtained by disjoint triplets of workers, which  achieves F1 of \textbf{84.1}, averaged over 4 experiments, 35 predicates each. 
Notably, consolidation boosts agreement, indicating its necessity.   
For \textsf{LA} agreement, averaged F1 was 67.8; however, it is likely that the drop from \textsf{UA} is mainly due to falsely rejecting semantically equivalent questions under the \textsc{Strict-Match} criterion, given that we found equal  \textsf{LA} and \textsf{UA} scores in a manual evaluation of our dataset (see Table~\ref{tab:expert_comparison} below).

\paragraph{Dataset Assessment and Comparison}
We assess our gold standard, as well as the recent {\sf Dense} set, against an integrated \textbf{expert set} of 100 predicates. 
To construct the expert set, we first merged the annotations from the {\sf Dense} set with our workers' annotations. 
Then, three of the authors blindly (i.e., without knowing the origin of each QA pair) selected, corrected and added annotations, resulting in a high-coverage unbiased expert set.
We further manually corrected the evaluation decisions, accounting for some automatic evaluation mistakes introduced by the span-matching and question equivalence criteria.
As seen in Table \ref{tab:expert_comparison}, 
our gold set yields comparable precision with drastically higher recall, in line with our 25\% higher yield.\footnote{The UA and LA measures ended up equal for our dataset after manual inspection since we found that all \emph{correctly} classified unlabeled arguments were annotated with a correct question role label.}
\begin{table}[ht]
\begin{center}
\resizebox{\columnwidth}{!}{
\begin{tabular}{cl|ccc|ccc}
\multicolumn{1}{l}{} &  & \multicolumn{3}{c|}{This work} & \multicolumn{3}{c}{Dense (2018)} \\
\multicolumn{1}{l}{} &  & \textbf{P} & \textbf{R} & \textbf{F1} & \textbf{P} & \textbf{R} & \textbf{F1} \\ \hline
\multicolumn{1}{c|}{\multirow{2}{*}{\bf UA}} & Auto. & 79.9 & 89.4 & 84.4 & 67.1 & 69.5 & 68.3 \\
\multicolumn{1}{c|}{} & Man. & \multicolumn{1}{l}{88.0} & \multicolumn{1}{l}{95.5} & \multicolumn{1}{l|}{91.6} & \multicolumn{1}{l}{86.4} & \multicolumn{1}{l}{70.5} & \multicolumn{1}{l}{77.6} \\ \hline
\multicolumn{1}{c|}{\multirow{2}{*}{\bf LA}} & Auto. & 71.0 & 79.5 & 75.0 & 49.5 & 51.3 & 50.4 \\
\multicolumn{1}{c|}{} & Man. & \multicolumn{1}{l}{88.0} & \multicolumn{1}{l}{95.5} & \multicolumn{1}{l|}{91.6} & \multicolumn{1}{l}{83.1} & \multicolumn{1}{l}{67.8} & \multicolumn{1}{l}{74.7} \\ \hline
\end{tabular}}
\caption{Automatic and manually-corrected evaluation of our gold standard and {\sf Dense} \cite{fitz2018qasrl} against the integrated expert    set.}
\label{tab:expert_comparison}
\end{center}
\end{table}

Examining disagreements between our gold and {\sf Dense}, we observe that our workers successfully produced more roles, both implicit and explicit. To a lesser extent, they split more arguments into independent answers, as emphasized by our guidelines, an issue that was left under-specified in previous annotation guidelines.
\paragraph{Agreement with PropBank Data}
\label{sec:propbank_agreement}
It is illuminating to observe the agreement between QA-SRL  and PropBank (CoNLL-2009) annotations \cite{hajivc2009conll}.
In Table~\ref{tab:propbank}, we replicate the experiments in \citet[Section 3.4]{He2015qasrl} for both our gold set and theirs, over a sample of 200 sentences from the Wall Street Journal (evaluation is automatic and the metric is similar to our \textsf{UA}). 
We report macro-averaged (over predicates) precision and recall for all roles, including core and adjuncts,\footnote{Core roles are A0-A5 in PropBank (recall) and QAs having {\it what} and {\it who} WH-words in QA-SRL (precision).} while considering the PropBank data as the reference set.
Our recall of PropBank roles is notably high, reconfirming the coverage obtained by our annotation protocol.

The measured precision with respect to PropBank is low for adjuncts, but this is due to the fact that QA-SRL captures many \textit{correct} implicit arguments, which fall out of PropBank's scope (where arguments are directly syntactically linked to the predicate). 
To examine this, we analyzed 100 arguments in our dataset not found in PropBank (``false positives"). 
We found that only 32 were due to wrong or incomplete QA annotations, while most others were valid implicit arguments, stressing QA-SRL's advantage in capturing those inherently. 
Extrapolating from this analysis estimates our true precision (on all roles) to be about 91\%, consistent with the 88\% precision in Table \ref{tab:expert_comparison}, while yielding about 15\% more valid arguments than PropBank (mostly implicit).
Compared with {\sf 2015}, our QA-SRL gold yielded 1593 QA pairs (of which, 604 adjuncts), 
while theirs yielded 1315 QAs (336 adjuncts). 
Overall, the comparison to PropBank reinforces the quality of our gold dataset and shows its better coverage relative to the \textsf{2015} dataset.
\begin{table}[ht]
\centering
\resizebox{\columnwidth}{!}{
\begin{tabular}{l|ccc|ccc}
& \multicolumn{3}{c|}{This work} & \multicolumn{3}{c}{{\citet{He2015qasrl}}} \\
 & \textbf{P} & \textbf{R}  & \textbf{F1}  & \textbf{P} & \textbf{R}  & \textbf{F1} \\ \hline

 \bf{All} &  73.3	& 93.0	& 82.0	& 81.7	& 86.6	& 84.1 \\
 \bf{Core} & 87.3	& 94.8	& 90.9	& 86.6	& 90.4	& 88.5 \\
 \bf{Adj.} & 43.4	& 85.9	& 57.7	& 59.7	& 64.7	& 62.1 \\ 
 \end{tabular}}
\caption{Performance analysis when considering PropBank as reference (all roles, core roles, and adjuncts).}
    \label{tab:propbank}
\end{table}

\section {Baseline Parser Evaluation} 
\label{sec:baseline_parser}
We evaluate the parser from \citet{fitz2018qasrl} on our dataset, providing a baseline for future work.
As we previously mention, unlike typical SRL systems, the parser outputs overlapping arguments, often with redundant roles (Table~\ref{tab:parser}). Hence, we employ our metric variant for evaluating redundant annotations. 
Results are reported in Table \ref{tab:pvg}, demonstrating reasonable performance along with substantial room for improvement, especially with respect to coverage. 
As expected, the parser's recall against our gold is substantially lower than the 84.2 recall reported in \cite{fitz2018qasrl} against {\sf Dense}, due to the limited recall of {\sf Dense} relative to our gold set.

\begin{table}[tb!]

\begin{center}
\resizebox{\columnwidth}{!}{
\begin{tabular}{l|ccc|ccc|ccc}
 & \multicolumn{3}{c|}{Test} & \multicolumn{6}{c}{Dev (\textit{Wikinews})} \\
 & \multicolumn{3}{c|}{Automatic} & \multicolumn{3}{c|}{Automatic} & \multicolumn{3}{c}{Manual} \\ 
 & P & R & F1 & P & R & F1 & P & R & F1\\ \hline
\textbf{UA} & 87.1 & 50.2 & 63.7 & 86.6                & 58.8       & 70.1 & 87.8       & 66.5       & 75.5  \\ \hline
\textbf{LA} & 67.8 & 39.1 & 49.6 & 65.0               & 44.2        & 52.6 & 83.9          & 64.3       & 72.8  \\\hline
\end{tabular}}
\caption{Automatic parser evaluation against our test set, complemented by automatic and manual evaluations on the Wikinews part of the dev set (manual evaluation is over 50 sampled predicates).}
\label{tab:pvg}
\end{center}
\end{table}
\paragraph{Error Analysis} 
Through manual evaluation of 50 sampled predicates, we detect correctly predicted arguments and questions that were rejected by the \textsc{IOU} and \textsc{Strict-Match} criteria.
Based on this inspection, out of the 154 gold roles (128 explicit and 26 implicit), the parser misses 23\%, covering 82\% of the explicit roles but only half of the implicit ones.

\section{Conclusion}
Applying our proposed controlled crowdsourcing protocol to QA-SRL successfully attains truly scalable high-quality annotation by laymen, facilitating future research of this paradigm.
Exploiting the open nature of the QA-SRL schema, our non-expert annotators produce rich argument sets with many valuable implicit arguments. 
Indeed, thanks to effective and practical training over the crowdsourcing platform, our workers' annotation quality, and particularly its coverage, are on par with expert annotation. 
We release our data, software and protocol, enabling easy future dataset production and evaluation for QA-SRL, as well as possible extensions of the QA-based semantic annotation paradigm.
Finally, we suggest that our simple yet rigorous controlled crowdsourcing protocol would be effective for other challenging annotation tasks, which often prove to be a hurdle for research projects.
\begin{table}[t!]
\resizebox{\columnwidth}{!}{
\begin{tabular}{ll}

What suggests something?& Reports\\
What suggests something? & Reports from Minnesota\\ 
\hline
Where was someone carried? & to reclining chairs \\
What was someone carried to? & reclining chairs \\
\end{tabular}}
\caption{Examples where \citet{fitz2018qasrl}'s parser generates redundant arguments. The first two rows illustrate different, partly redundant, argument spans for the same question, while the bottom rows illustrate two paraphrased questions for the same role.}
\label{tab:parser}
\end{table}

\section*{Acknowledgments}
\vspace{-5pt}
This work was supported in part by an Intel Labs grant, the Israel Science Foundation grant 1951/17 and the German Research Foundation through the German-Israeli Project Cooperation (DIP, grant DA 1600/1-1).

\bibliography{000_ACL20_qasrl_ground_truth}
\bibliographystyle{acl_natbib}

\appendix

\section{Appendix}
\label{sec:supplemental}

\paragraph{Evaluating Redundant Annotations}
Recent datasets and parser outputs of QA-SRL (Fitzgerald et al., 2018)  produce redundant arguments. On the other hand, our consolidated gold data, as typical, consists of a single non-redundant annotation, where arguments are non-overlapping.
In order to fairly evaluate such redundant annotations against our gold standard, 
we ignore predicted arguments that match ground-truth but are not selected by the bipartite matching due to redundancy. 
After connecting unmatched predicted arguments that overlap, we count one false positive for every connected component, aiming to avoid penalizing precision too harshly when predictions are redundant.

\end{document}